\documentclass[letterpaper, 10 pt, conference]{ieeeconf}  

\IEEEoverridecommandlockouts                              

\usepackage{amsmath} 
\usepackage[nolist,nohyperlinks]{acronym}
\usepackage{todonotes}
\usepackage{graphicx}
\usepackage{subcaption}
\usepackage{amsfonts} 
\usepackage{float}
\usepackage{comment}
\usepackage{siunitx}
\usepackage{adjustbox}
\usepackage{array}
\usepackage{booktabs} 
\usepackage{hyperref}
\usepackage{fontawesome5}
\usepackage{wrapfig}
\usepackage{lipsum}
\usepackage{xspace}
\usepackage{tabularx}

\usepackage[ruled]{algorithm2e}

\SetKwComment{Comment}{/* }{ */}

\acrodef{AI}{Artificial Intelligence}
\acrodef{CAD}{Computer-Aided Design}
\acrodef{NeRF}{Neural Radiance Field}
\acrodef{NGP}{Neural Graphics Primitives}
\acrodef{DoF}{Degrees of Freedom}
\acrodef{SDF}{Signed Distance Field}
\acrodef{MLP}{Multi-Layer Perceptron}

\def\eg{\emph{e.g.,}\xspace} 

\def\ie{\emph{i.e.,}\xspace}

\def\aka{\emph{a.k.a.}\xspace}

\definecolor{lightgreen}{HTML}{CEEAD6}
\definecolor{lightred}{HTML}{FAD2CF}
\definecolor{lightorange}{HTML}{FEEFC3}
\definecolor{lightblue}{HTML}{30CEFE}
\definecolor{darkerblue}{HTML}{5CA3FF}
\definecolor{brightpurple}{HTML}{9865FE}

\newcommand{\ignore}[1]{}

\newcommand{\ba}{\begin{array}}
\newcommand{\ea}{\end{array}}
\newcommand{\bc}{\begin{center}}
\newcommand{\ec}{\end{center}}
\newcommand{\be}{\begin{enumerate}}
\newcommand{\ee}{\end{enumerate}}
\newcommand{\bea}{\begin{eqnarray}}
\newcommand{\eea}{\end{eqnarray}}
\newcommand{\beas}{\begin{eqnarray*}}
\newcommand{\eeas}{\end{eqnarray*}}
\newcommand{\beq}{\begin{equation}}
\newcommand{\eeq}{\end{equation}}
\newcommand{\bfig}{\begin{figure}}
\newcommand{\efig}{\end{figure}}
\newcommand{\bi}{\begin{itemize}}
\newcommand{\ei}{\end{itemize}}
\newcommand{\bpic}{\begin{picture}}
\newcommand{\epic}{\end{picture}}
\newcommand{\btabular}{\begin{tabular}}
\newcommand{\etabular}{\end{tabular}}
\newcommand{\btable}{\begin{table}}
\newcommand{\etable}{\end{table}}

\newcommand{\es}{\vfill
                 \rule[-6mm]{170mm}{0.7mm} \\
                 \redw{{\tiny
		  \hfill S-\theslide}}
                 \end{slide}}

\def \hbar {{\bar{h}}}

\makeatletter
\renewcommand*\env@matrix[1][*\c@MaxMatrixCols c]{%
  \hskip -\arraycolsep
  \let\@ifnextchar\new@ifnextchar
  \array{#1}}
\makeatother

\title{\LARGE \bf
\emph{Attentive Feature Aggregation}\\ or: How Policies Learn to Stop Worrying about Robustness and \\Attend to Task-Relevant Visual Cues$^\dagger$ 
} 

\author{\ Nikolaos Tsagkas$^{1,\alpha}$,\ \ Andreas Sochopoulos$^1$, \ Duolikun Danier$^1$,\\\ Sethu Vijayakumar$^1$, Alexandros Kouris$^3$, \ Oisin Mac Aodha$^{1,\varepsilon}$,\ Chris Xiaoxuan Lu$^{2, \varepsilon}$\\ %
  $^1$University of Edinburgh, \ $^2$UCL, \ $^3$Samsung AI Center - Cambridge, UK\\
\thanks{*This work was supported by the United Kingdom Research and Innovation (grant EP/S023208/1), EPSRC Centre for Doctoral Training in Robotics and Autonomous Systems (RAS) and ELIAI (Edinburgh Laboratory for Integrated Artificial Intelligence) - EPSRC (EP/W002876/1).\newline
\indent$^\dagger$The title is inspired by Stanley Kubrick's film:~\textit{``Dr. Strangelove or: How I Learned to Stop Worrying and Love the Bomb''.}\newline
\indent$^\varepsilon$ Indicates equal senior authorship.\newline
\indent$^\alpha$ Corresponding author: N.~Tsagkas -- n.tsagkas@ed.ac.uk}
}

\begin{document}

\maketitle
\thispagestyle{empty}
\pagestyle{empty}


\begin{abstract}
The adoption of pre-trained visual representations (PVRs), leveraging features from large-scale vision models, has become a popular paradigm for training visuomotor policies. 
However, these powerful representations can encode a broad range of task-irrelevant scene information, making the resulting trained policies vulnerable to out-of-domain visual changes and distractors. 
In this work, we investigate visuomotor policy feature pooling as a solution to the observed lack of robustness in perturbed scenes.
We achieve this via Attentive Feature Aggregation (AFA), a lightweight, trainable pooling mechanism that learns to naturally attend to task-relevant visual cues while ignoring semantically rich scene distractors.   
Through extensive experiments in both simulation and the real world, we demonstrate that policies trained with AFA significantly outperform standard pooling approaches in the presence of visual perturbations, without requiring expensive dataset augmentation or fine-tuning of the PVR. 
Our findings show that ignoring extraneous visual information is a crucial step towards deploying robust and generalisable visuomotor policies.\newline
Project Page:~\texttt{\href{https://tsagkas.github.io/afa}{tsagkas.github.io/afa}}.\newline
\end{abstract}

\section{INTRODUCTION}
Performing robust and accurate robotic manipulation from visual inputs necessitates informative and stable visual representations. 
The traditional paradigm for training visuomotor policies has involved learning visual encoders from scratch alongside policy models~\cite{JMLR:v17:15-522}. 
Recently, however, the adoption of pre-trained visual representations (PVRs) (\aka vision foundation models), \ie computer vision models trained on large and diverse visual datasets, has emerged as a compelling alternative, moving away from the tabula-rasa approach~\cite{shang2024theia,jing2023explore,NEURIPS2023_022ca1be,silwal2024large,dasari2023datasets,hsu2022what}. 
This shift has been driven mainly by three key factors: the SoTA performance of PVRs in computer vision tasks, their impressive generalisation capabilities derived from training on vast datasets, and the absence of \emph{robust} robot-specific foundation models~\cite{huang2025otter}.

\begin{figure}[t]
    \centering
    \includegraphics[width=1\linewidth]{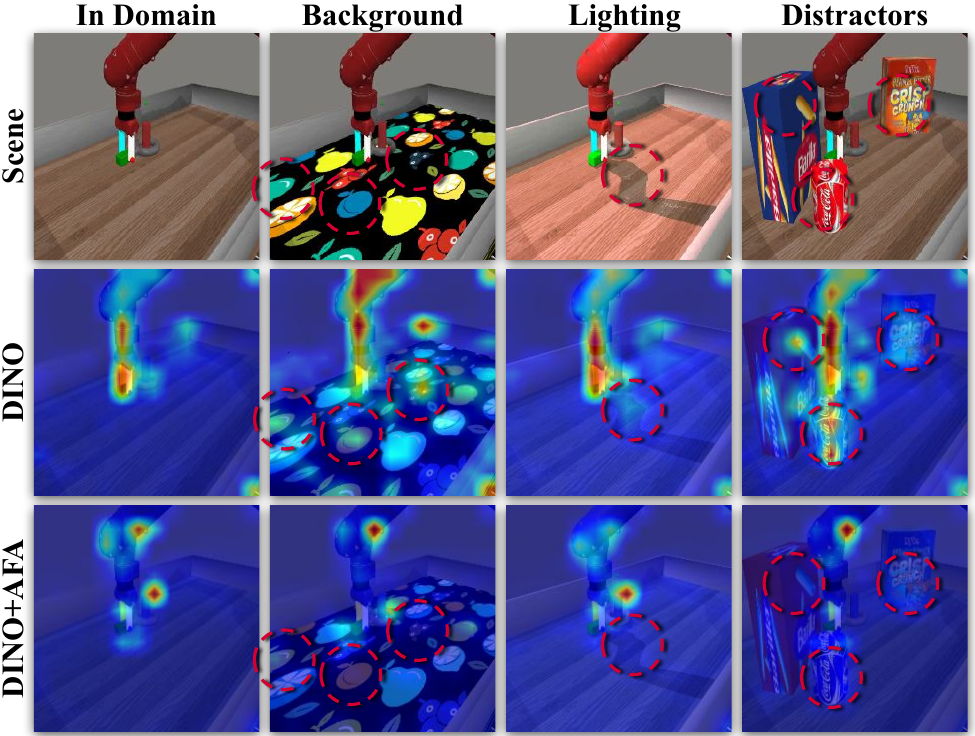}
    \caption{Comparison of attention heatmaps with and without our AFA approach. AFA learns to attend to focused, task-relevant regions, ignoring scene changes (e.g., distractors).}
    \label{fig:teaser}
    \vspace{-0.99em}  
\end{figure}

Even though PVRs have played an instrumental role in downstream robotics applications, such as semantic mapping~\cite{tsagkas2023vlfields,tsagkas2024click}, language-guided~\cite{shen2023F3RM,huang2023voxposer} and affordance-guided~\cite{ju2025robo,li2024affgrasp} manipulation, their adoption in imitation learning, although offering unprecedented data efficiency in policy training, can lead to a paradoxical phenomenon. 
The very PVRs that are selected for their ability to provide descriptive and generalisable features, also encode a broad range of scene information, much of which may be irrelevant to the specific task.
As a result, introducing visual changes in the scene, particularly ones with rich semantic meaning, as in Fig.~\ref{fig:teaser}, the more likely it is for the policy's input to be driven out-of-domain (OOD), leading to task failure. 
This lack of robustness is a well-known problem~\cite{10611331,Hansen2022pre,burns2024what,houlsby2019parameter,spawnet}. 
Nevertheless, prior works either simply acknowledged and quantified the problem, or proposed solutions that required dataset augmentation~\cite{Hansen2022pre,spawnet} (\eg randomising the background with distractor videos), a costly approach for real-world robotics applications.   

In this work, we argue that an effective way of improving a policy's robustness should leave the PVR out of the fine-tuning loop, which ideally should be kept frozen to avoid diluting the representation. 
Similarly, the methodology is better to not rely on dataset augmentation via domain randomisation, which is difficult and costly to run in real-world imitation learning.
At the same time, dataset enhancements in-simulation require dealing with \texttt{sim2real} transfer, a problem that is not trivial to solve. 
In contrast, we propose a solution that focuses on the way PVR feature extraction is handled by learning to attend only to task-specific information.

In summary, we make the following contributions:\\
\textbf{1.~Re-thinking visuomotor policy feature pooling}. We propose the use of a trainable module, inspired by recent work in computer vision~\cite{Chen2024, danier2024depthcuesevaluatingmonoculardepth, bardes2024revisiting}, called \textit{Attentive Feature Aggregation (AFA)}, for the purpose of increasing robustness under scene perturbations which  naturally learns to attend to task-relevant visual cues.
Our method outperforms greatly standard approaches, while in some cases even triples OOD performance.\\
\textbf{2.~Introduce robustness predictors}.
We evaluate the use of AFA both in in-domain (ID) and OOD scenes, under lighting and background visual perturbations. 
We validate its superior robustness  across 14 popular PVRs and two SoTA feature pooling approaches, widely adopted in visuomotor policy learning.
We further verify AFA's efficacy in the real-world, on two different robot platforms.
From these results, we make the observation that attention-related metrics provide insight into OOD performance. 
More specifically, we find that the amount of attention mass that falls within task-relevant regions and the attention entropy score (\ie how targetted the attention is) are strongly correlated with the OOD policy performance.
AFA improves both metrics, which justifies its increased performance.

\section{RELATED WORK}
\noindent \textbf{Robustness in PVR-based policies}: PVRs are favoured for their generalisation capabilities in vision tasks, but OOD generalisation remains challenging in policy deployment. 
\cite{10611331} analysed the impact of various perturbations on PVR-based policy generalisation, while~\cite{burns2024what} identified correlations between generalisation and inherent model traits, such as ViTs' segmentation ability.
Conversely,~\cite{Hansen2022pre} found that learning from scratch with data augmentation can yield competitive results, while~\cite{spawnet} found that adapters~\cite{houlsby2019parameter} can improve policy generalisation when training with diverse object instances. 
In contrast, we focus on developing methods that achieve robustness to scene changes without relying on dataset augmentation, which can be prohibitively expensive in real-world robotics applications, or fine-tuning of the PVR, which could harm their generalisation properties.

\noindent \textbf{Isolating task-relevant features}: Downstream vision tasks often make use of the output features of PVRs (\eg semantic correspondence~\cite{Mariotti_2024_CVPR}, depth estimation~\cite{NEURIPS2024_26cfdcd8}, etc.). 
However, these features typically encode a broad range of scene information, much of which may be irrelevant to the specific task. 
To address this challenge, attentive probing~\cite{Chen2024, danier2024depthcuesevaluatingmonoculardepth, bardes2024revisiting} has emerged as a popular evaluation technique, leveraging local tokens. 
This approach employs a cross-attention layer with a trainable query token, treating the local features from PVRs as a sequence of key-value pairs. 
Unlike traditional evaluation methods such as linear probing (\ie stacking a linear layer after a pre-trained model to evaluate the capabilities of the output feature on a simple task), attentive probing (\ie deploying a cross-attention layer with a trainable query token to process the local output features), has shown significantly different vision evaluation outcomes, particularly with PVRs trained using Masked-Image Modelling (MIM) approaches (\eg MAE~\cite{He2021MaskedAA}), where features, such as the CLS token, often include irrelevant information.
Similarly, in robot learning, task-relevant signals like joint angles may correspond to particular image regions, with unrelated cues acting as distractions. 
In this work, we find that attentive probing is key to robust and generalizable visuomotor policy learning too, where the trainable query token learns to attend to task-relevant visual cues.

\noindent \textbf{Policy observation pooling}: In robotics applications, the idea of pooling a sequence of observation tokens has been explored, and it is usually deployed for reducing the input stream length. 
A popular choice remains to this day adding a Spatial Softmax operation between the visual encoder and the policy network~\cite{he2016deep,chi2023diffusionpolicy}.
Similarly, RT-1~\cite{brohan2023rt1roboticstransformerrealworld} utilised a TokenLearner~\cite{ryoo2021tokenlearner} for making the policy input more compact, thus speeding up inference time. 
However, there is no current consensus as to how PVR features can be effectively used such that the policy model learns to attend only to task-relevant signals, ignoring unrelated cues that act as distractors.
By prioritizing task-relevant signals, we show that \textit{Attentive Feature Aggregation (AFA)} enhances task performance in OOD scenarios, outperforming traditional pooling methods. 
Recently, PerAct~\cite{shridhar2022peract} also adopted attentive probing in the context of visuomotor policy learning, but for the purpose of summarising long token sequences from voxel-based spatial representations, to enable their efficient processing by subsequent self-attention layers.  
Also, ICRT~\cite{fu2024icrt} followed a similar methodology, but for the purpose of learning to project image observations from different views, the robot's proprioception and actions into a joint latent space for conducting sequence modelling.

To the best of our knowledge, we are the first to utilise attentive probing for increasing the policy's robustness.

\section{PRELIMINARIES}
\subsection{Imitation Learning via Behaviour Cloning}
\label{ssec:method_preliminaries}

We assume access to an expert policy $\pi^\star$ that prescribes robot actions based on both proprioceptive input $p \in \mathcal{P}$ and visual input $o \in \mathcal{O}$. 
The output is an action $a \in \mathcal{A}$. 
Executing this controller yields a collection of expert rollouts $\mathcal{T}^\text{e} = \{(p_t^i, o_t^i, a_t^i)_{t=0}^T\}_{i=1}^N$,
where each of the $N$ trajectories is composed of $T$ time steps of observations paired with corresponding actions.  

\begin{figure}[t]
    \centering
    \includegraphics[width=0.95\linewidth]{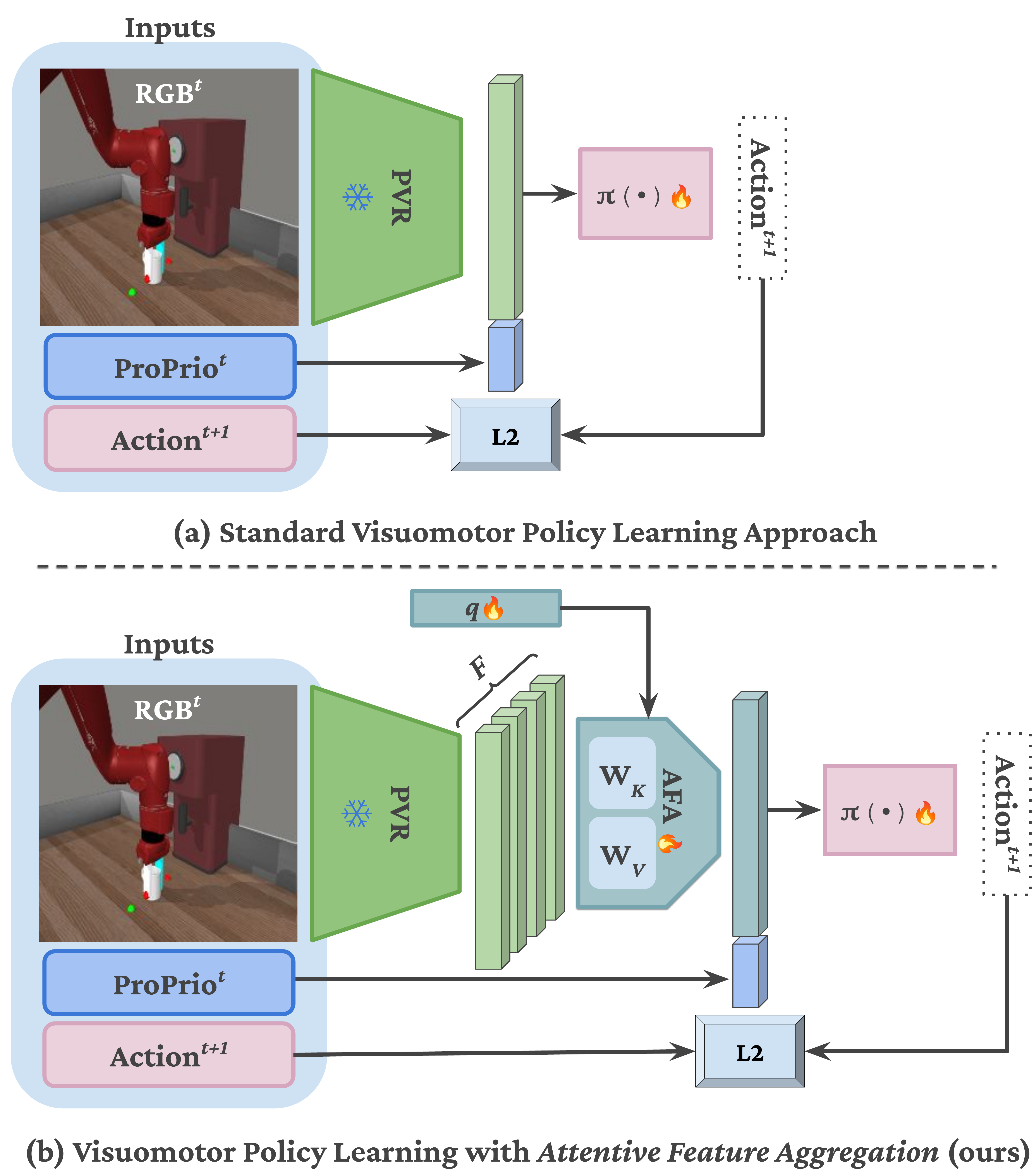}
    \caption{Visualisation of (a) the standard visuomotor policy learning approach and (b) our proposed AFA approach.}
    \label{fig:policy}
    \vspace{-15pt}
\end{figure}

The learner's objective is to approximate $\pi^\star$ with a parameterised policy $\pi_\theta$. 
We adopt standard behaviour cloning, in which $\pi_\theta$ is trained on expert demonstrations by minimizing the error between expert actions and the policy's predictions:
\begin{equation}
   \mathbb{E}_{(p_t^i, o_t^i, a_t^i) \sim \mathcal{T}^\text{e}} 
   \big\| a_t^i - \pi_\theta(f_\text{PVR}(o_t^i),\, p_t^i) \big\|_2^2,
\end{equation}
where the function $f_\text{PVR}$ denotes a frozen PVR that transforms the raw visual stream $o_t^i$ into global feature vectors (see Fig.~\ref{fig:policy}a).  

Following common practice~\cite{parisi2022unsurprising, nair2022rm, hu2023pretrainedvisionmodelsmotor}, we implement $\pi_\theta$ as a lightweight multilayer perceptron, with 4 layers, separated by ReLUs, with a tanh activation at the end to constrain outputs. 
The policy predicts the mean $\mu$ of a Gaussian distribution with a fixed variance $\sigma^2$, and actions are sampled from a truncated Gaussian $\mathcal{N^T}(\mu, \sigma^2)$ restricted to the range $[-1, 1]^k$, where $k$ is the dimensionality of the action space.  

\subsection{Policy Pooling Mechanisms}
\noindent\textbf{Spatial Softmax} has been widely used in visuomotor policy learning~\cite{finn2016deepspatialautoencodersvisuomotor,chi2023diffusionpolicy} for extracting spatially meaningful, compressed features from convolutional feature maps. 
This pooling method operates by interpreting the activation map as a probability distribution over spatial locations, thereby encouraging the network to focus on the most feature-rich regions in the image. 
This results in a differentiable way to extract expected spatial coordinates of visual features, which can be directly fed into downstream policy networks.

Given a feature map \( F \in \mathbb{R}^{H \times W \times D} \), where \( D \) is the number of channels, and \( H \), \( W \) are the height and width respectively, the Spatial Softmax is applied to each channel \( d \) as follows:

\begin{equation}
s_{ij}^{(d)} = \frac{\exp(f_{ij}^{(d)})}{\sum\limits_{i'=1}^{H} \sum\limits_{j'=1}^{W} \exp(f_{i'j'}^{(d)}).}
\end{equation}

\noindent Using the resulting softmax weights \( s_{ij}^{(d)} \), the expected 2D coordinates \( (\hat{x}^{(d)}, \hat{y}^{(d)}) \) for each channel are computed as: $\hat{x}^{(d)} = \sum_{i=1}^{H} \sum_{j=1}^{W} s_{ij}^{(d)} \cdot x_j$, $\hat{y}^{(d)} = \sum_{i=1}^{H} \sum_{j=1}^{W} s_{ij}^{(d)} \cdot y_i$, where \( x_j \) and \( y_i \) denote the horizontal and vertical coordinates normalised to a fixed range (\ie \([-1, 1]\)).  


This pooling technique thus compresses high-dimensional spatial information into a compact and interpretable form, reducing the amount of input information to the policy model, while remaining end-to-end differentiable.
Nevertheless, this does not guarantee that the the pooling method isolates the task-relevant info, nor that it will ignore strong visual cues that may act as distractors in OOD scenes. 
Indeed, we find that Spatial Softmax struggles to perform in OOD scenarios, even though it leads to a slight increase in ID scenes. 

\noindent\textbf{TokenLearner}~\cite{ryoo2021tokenlearner} reduces computational complexity by dynamically abstracting visual inputs into a small set of informative tokens. 
Unlike static pooling operations, TokenLearner (TL) introduces learnable spatial attention maps, conditioned on the input and trained jointly with the backbone, 
allowing the model to adaptively highlight regions of interest. 
Due to its efficiency in compressing high-dimensional data, it has been adopted in robotic control architectures such as RT-1~\cite{brohan2023rt1roboticstransformerrealworld} to accelerate inference.

Given an input feature map \( F \in \mathbb{R}^{H \times W \times D} \), TL computes \( M \) spatial attention maps \( A^{(m)} \in \mathbb{R}^{H \times W} \). 
These maps are generated via a lightweight function \( \phi_m \), parameterised by convolutional layers, followed by a sigmoid ($\sigma$) activation function $A^{(m)} = \sigma(\phi_m(F)), \quad \text{for } m = 1, \dots, M$. 
Crucially, each learned token \( T^{(m)} \in \mathbb{R}^{D} \) is computed by applying \textit{spatial global average pooling} to the element-wise product of the input features and the attention map:

\begin{equation}
T^{(m)} = \frac{1}{H \times W} \sum_{i=1}^{H} \sum_{j=1}^{W} A_{ij}^{(m)} \odot F_{ij}.
\end{equation}

The resulting set of tokens \( \{T^{(1)}, \dots, T^{(M)}\} \in \mathbb{R}^{M \times D} \) provide a compact semantic summary of the scene. 
In our framework, we compare AFA to TL by evaluating their capacity to isolate and preserve task-relevant visual cues, in the context of frozen PVR-based policy learning.  
While both methods introduce adaptive pooling, we 
argue that TL can cause spatial information loss affecting the policy's performance in both ID and OOD settings, in contrast to AFA that effectively encourages robustness and generalisation under visual scene perturbations. 

\section{METHODOLOGY}
We posit that training policies using the global features of PVRs (\ie CLS token for ViTs or average pooled channel feature for CNNs) can lead to overfitting to scene conditions that are irrelevant to the task at hand. 
The output features of these representations often capture visual characteristics of the scene that may be irrelevant to the policy (\eg the texture of a tabletop). 
Processing such extraneous information not only dilutes the policy network's focus, but also leads to overfitting to  scene specific conditions. 
This observation aligns with recent work on vision model evaluation~\cite{Chen2024}, which argues that only specific image regions carry the necessary information for solving a task.

\begin{table}[t]
\vspace{1.5em} 
\centering
\caption{Summary of evaluated PVRs. \faRobot~denotes models trained for robotics. Size is number of images, unless specified. Datasets: \textbf{IN}: ImageNet~\cite{ridnik2021imagenet21kpretrainingmasses}, \textbf{LVD}: LVD-142M~\cite{oquab2023dinov2}, \textbf{K}: Kinetics~\cite{kay2017kineticshumanactionvideo}, \textbf{E4D}: Ego4D~\cite{grauman2022ego4dworld3000hours}, and \textbf{E4D+MNI}~\cite{NEURIPS2023_022ca1be}.}
\begin{adjustbox}{max width=\linewidth}
%
\label{tab:pvrs_info}
\setlength{\tabcolsep}{3pt} 
\footnotesize 
\renewcommand{\arraystretch}{1.1} 
\begin{tabular}{@{}l l l l  p{2.3cm}@{}} 
\toprule
\textbf{Model} && \textbf{Arch.} & \textbf{Objective} & \textbf{Dataset (Size)}  \\
\midrule
MAE~\cite{He2021MaskedAA}     && ViT-B/16 & MIM              & IN (1.2M)  \\
VC-1~\cite{NEURIPS2023_022ca1be} &\faRobot & ViT-B/16 & MIM              & E4D+MNI (5.6M$^\dagger$) \\
DINOv1~\cite{caron2021emerging} && ViT-B/16 & Self-Distillation     & IN (1.2M)                   \\
iBOT~\cite{zhou2021ibot}     && ViT-B/16 & MIM+Self-Distillation & IN (14M)                   \\
DINOv2~\cite{oquab2023dinov2} && ViT-B/14 & MIM+Self-Distillation & LVD (142M)                 \\
ViT~\cite{dosovitskiy2021an}  && ViT-B/16 & Supervised       & IN (14M)                  \\
CLIP~\cite{radford2021learning}  && ViT-B/16 & V-L Contrastive  & LAION (2B)              \\
\midrule
MoCov2~\cite{chen2020mocov2}  && R-50     & Contrastive      & IN (1.2M)                   \\
DenseCL~\cite{wang2021dense} && R-50     & Local Contrastive & IN (1.2M)                   \\
SwAV~\cite{caron2020unsupervised} && R-50 & Clustering       & IN (1.2M)                   \\
VICRegL~\cite{bardes2022vicregl} && R-50 & VICReg (global+local)   & IN (1.2M)                    \\
VFS~\cite{xu2021rethinking}    && R-50     & Self-Distillation (video) & K (240K$^*$)            \\
VIP~\cite{ma2022vip} &\faRobot   & R-50     & Value Function    & E4D (5M$^\dagger$)       \\
R3M~\cite{nair2022rm} &\faRobot  &R-50     & Time Contrastive+Language     & E4D (5M$^\dagger$)       \\
\bottomrule
\multicolumn{5}{l}{\scriptsize $^\dagger$Number of frames from videos. $^*$Number of videos.}
\end{tabular}
\end{adjustbox}
\vspace{-2em}  
\end{table}

Building on this insight, we hypothesise that incorporating local information is particularly effective in the context of robot learning, echoing findings in PVR distillation research~\cite{shang2024theia}, though this area remains empirically under-explored, particularly since token pooling methods are utilised for reducing the computational load, instead of filtering out irrelevant cues that might drive the policy OOD.

Recognising the importance of local information is only part of the solution.
A data-driven mechanism is also required to filter irrelevant details, such as background patches, and prioritise task-relevant information. 
To this end, we adopt the \emph{attentive probing} methodology~\cite{Chen2024} to the context of visuomotor policy learning to implement \emph{Attentive Feature Aggregation} (AFA). 
Specifically, we append a cross-attention layer to the frozen PVR, modified to include a trainable query token that interacts with the sequence of local tokens produced by the model. 
This token implicitly learns to \textit{ask} the questions: \textit{``where do I need to look to solve the task?''}.
The local tokens correspond to the per-patch embeddings for ViTs and the channel embeddings for CNNs, both from the final layer. 
\begin{equation}
    \text{\textit{AFA}}(\boldsymbol{q}, F) = \text{\textit{softmax}}\left(\frac{\boldsymbol{q} \cdot (F \cdot W_K)^\top}{\sqrt{d_k}}\right) F \cdot W_V.
\end{equation}
The trainable query token $\boldsymbol{q}$ computes dot products with the feature sequence, with length equal to the number of patches and dimension $d_k$, organised as a matrix $F$. 
These dot products are passed through a softmax function to assign weights to the contributions of each local token to the final embedding. 
Our AFA module consists of multiple heads, so that specific dimension groups that might be irrelevant to the policy can be filtered out. 
Gradients are allowed to flow through the cross-attention layer, updating the parameters of $\boldsymbol{q}$ as well as the key and value projection matrices, $W_K$ and $W_V$. 
A visualisation of this approach is depicted in Fig.~\ref{fig:policy}b.

\begin{figure}[ht]
    \vspace{1.5em} 
    \centering
    \includegraphics[width=0.95\linewidth]{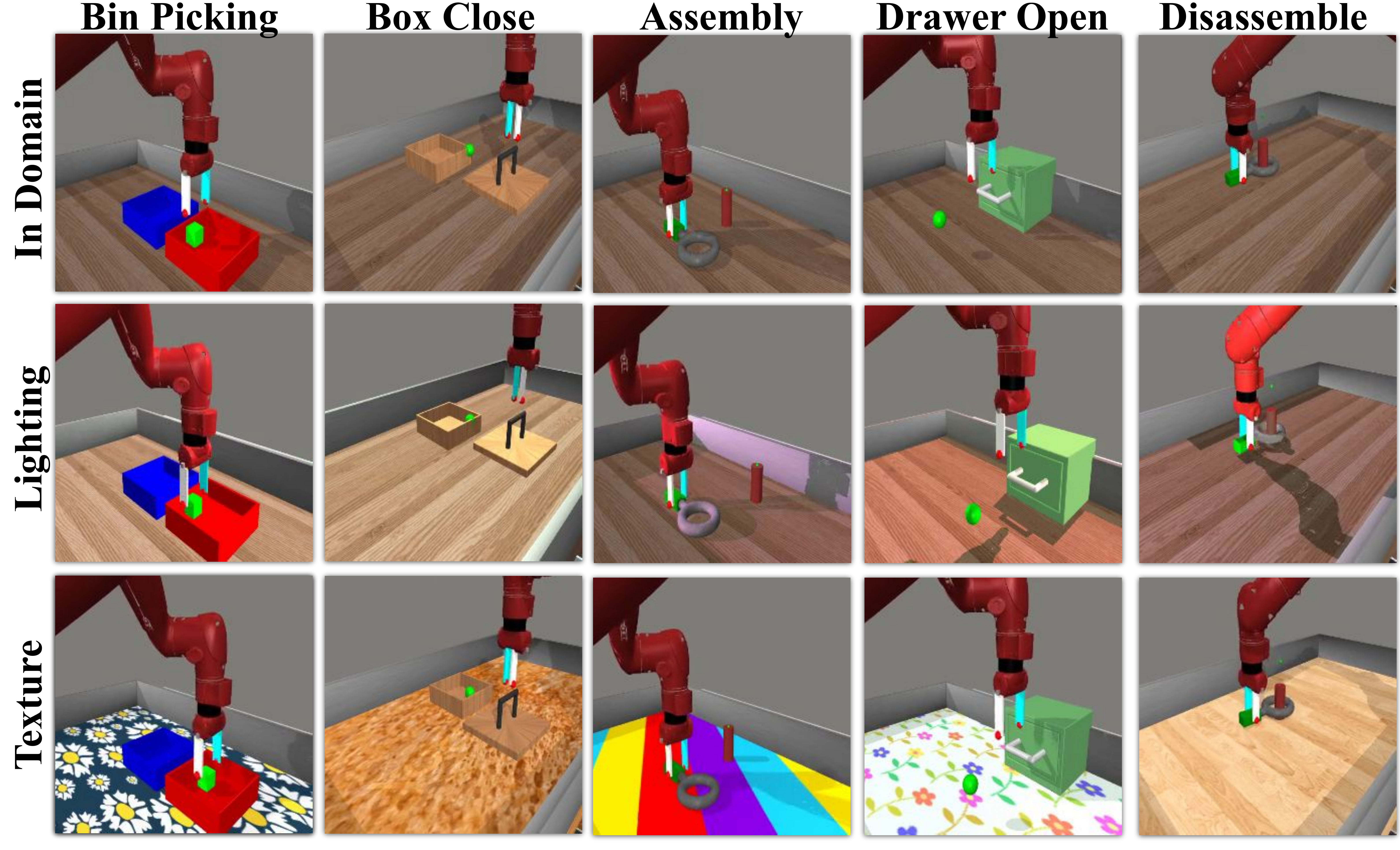}
    \caption{\textbf{Scene perturbations.} Top: Representative in-domain scenes. Middle: Randomised lighting (brightness, orientation, position). Bottom: Tabletop texture variations.}
    \label{fig:tasks}
    \vspace{-2em}  
\end{figure}

\section{EXPERIMENTS}
In this section, we introduce our experiments. 
First, in Section~\ref{ssec:exp_details} we summarise our implementation details, regarding the used environment, choice of PVRs, policy training approach and AFA hyperparameters. 
Second, in Section~\ref{ssec:robustness_eval} we analyse the ID and OOD performance of all PVRs raw and combined with different pooling methods.
Third, in Section~\ref{sec:ood_predictors}, we rely on our insight from the robustness evaluation to introduce two metrics that act as strong OOD policy performance predictors.
Finally, in Section~\ref{sec:real_world}, we validate our findings in the real-world. 

\subsection{Implementation Details}
\label{ssec:exp_details}
\noindent\textbf{Environment}. We conduct our experiments in the popular, MuJoCo-based~\cite{6386109} MetaWorld simulation environment~\cite{yu2020meta}. 
We select ten tasks (namely: {{box-close-v2}, {disassemble-v2}, {shelf-place-v2}, {peg-insert-side-v2}, {bin-picking-v2}, {coffee-pull-v2}, {assembly-v2}, {drawer-open-v2}, {pick-place-wall-v2}, and {button-press-wall-v2}}) and generate 25 expert demonstrations with a maximum of 175 rollout steps for each using the provided heuristic policies.

\begin{figure*}[t]
    \centering
    \vspace{1.5em} 
    \includegraphics[clip, width=0.95\textwidth]{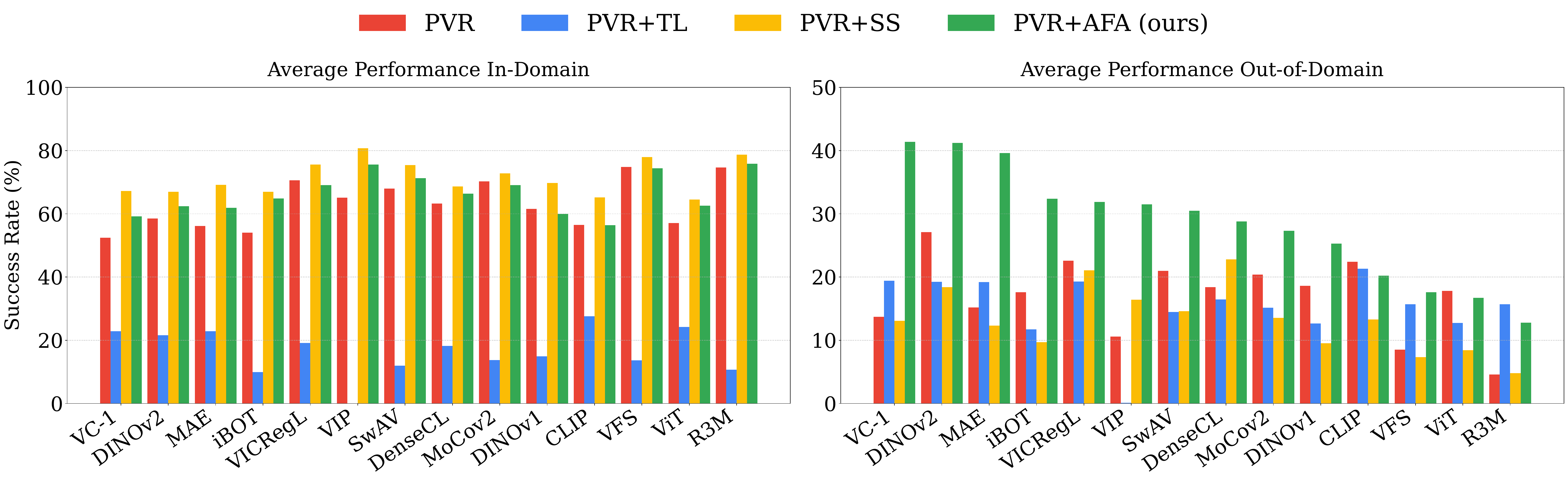}
    \caption{Success rate (\%) of policies trained with features from 14 PVRs in and out-of-domain. For \textbf{PVR}, the raw output PVR features are utilised (\ie \texttt{CLS} token for ViTs and the channel average for ResNets). \textbf{PVR+TL}, \textbf{PVR+SS}, and \textbf{PVR+AFA} stack a \textit{TokenLearner}, a Spatial Softmax, and an \textit{Attentive Feature Aggregation} pooling module after the PVR, respectively.}
    \label{fig:results}
    \vspace{-1.5em} 
\end{figure*}

\textit{Randomising the scene's lighting properties}. 
The brightness of the scene is altered by adjusting the diffuse light components, where each colour channel (red, green, blue) is randomly set to a value between 0.3 and 1.0. 
The specular highlights are similarly randomised, with lower intensity values ranging from 0.1 to 0.5. 
Additionally, the position of each light source is varied randomly within a 3D space, spanning horizontal and vertical shifts between -2 and 2 units and height adjustments between 0.5 and 3 units.
Lastly, the direction of the lights is randomised, allowing for changes in their angular orientation, with each directional component varying between -1 and 1 for horizontal/vertical angles and up to -1 for downward angles.

\textit{Randomising the scene's background}. 
We randomly modify the tabletop texture selecting from $30$ distinct textures (some of which are taken from~\cite{10611331}, visualised in the supplementary material).
Some are visually similar to the texture used in the training demos and others are vibrant, with patterns that hold semantic information that could potentially attract the attention of a PVR. 
Nevertheless, by observing the evaluation rollouts, policies can fail and succeed in both out-of-distribution cases.  
In Fig.~\ref{fig:tasks} we visualise some of the tasks and their corresponding visually perturbed variants. 

\noindent\textbf{PVRs}. To validate our hypotheses, we adopt seven Residual Networks (ResNets)~\cite{he2016deep} and seven Vision Transformers (ViT)~\cite{dosovitskiy2021an} as vision encoders for our policy, summarised in Tab.~\ref{tab:pvrs_info}. 
Our selection includes the most popular PVRs utilised in robot learning applications that have led to SoTA performance. 
We also aim to ensure diversity across training strategies, datasets, and the balance between local and global perception. 
The models tested include powerful representations from vision-specific approaches (\eg DINOv2~\cite{oquab2023dinov2}), vision-language models (\eg CLIP~\cite{radford2021learning}), and robot-learning-focused models (\eg R3M~\cite{nair2022rm}).
Despite these variations, we maintain a consistent backbone architecture of ResNet-50 or ViT-B/16, with the exception of DINOv2~\cite{oquab2023dinov2}, which employs a smaller patch size of 14. 
Also, for DINOv2 we discard overlapping patches, to ensure fairness in the comparison of PVRs. 

\noindent\textbf{Policy training}. We repeat each policy training five times using different seeds, keeping the PVR frozen, and report the interquartile  mean (IQM) success rate.
We train with mini-batches of 128 samples for 80K steps. 

\noindent\textbf{AFA}. We use a cross-attention layer with 12 heads for ViT features and 32 heads for ResNet features.
This configuration ensures that we process 64-dimensional feature chunks in both cases, maintaining fairness between the two backbone architectures. 
Consistent with standard attentive probing methodologies~\cite{Chen2024, danier2024depthcuesevaluatingmonoculardepth}, we employ a cosine learning rate scheduler with a warm-up phase. 

\begin{figure*}[t]
\vspace{1.5em} 
    \centering
    \begin{subfigure}[b]{0.48\textwidth} 
        \centering
        \includegraphics[width=0.95\linewidth, clip]{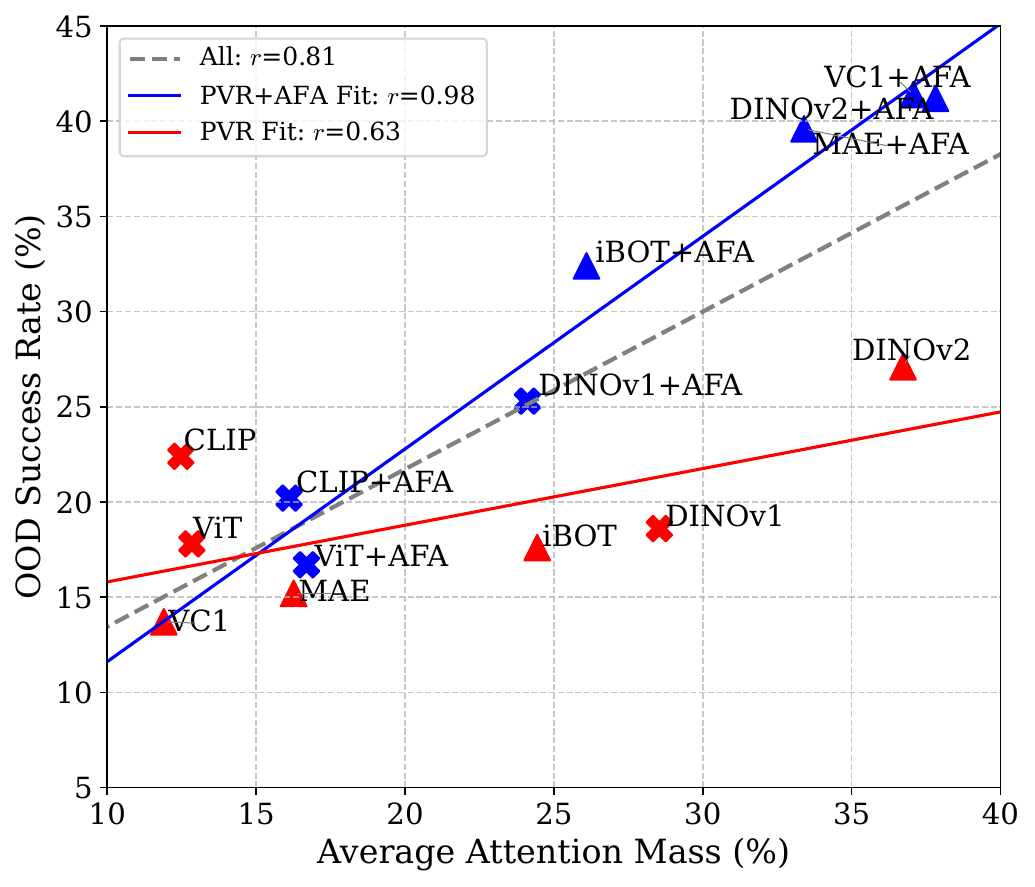}
    \end{subfigure}
    \begin{subfigure}[b]{0.48\textwidth} 
        \centering
        \includegraphics[width=0.95\linewidth, clip]{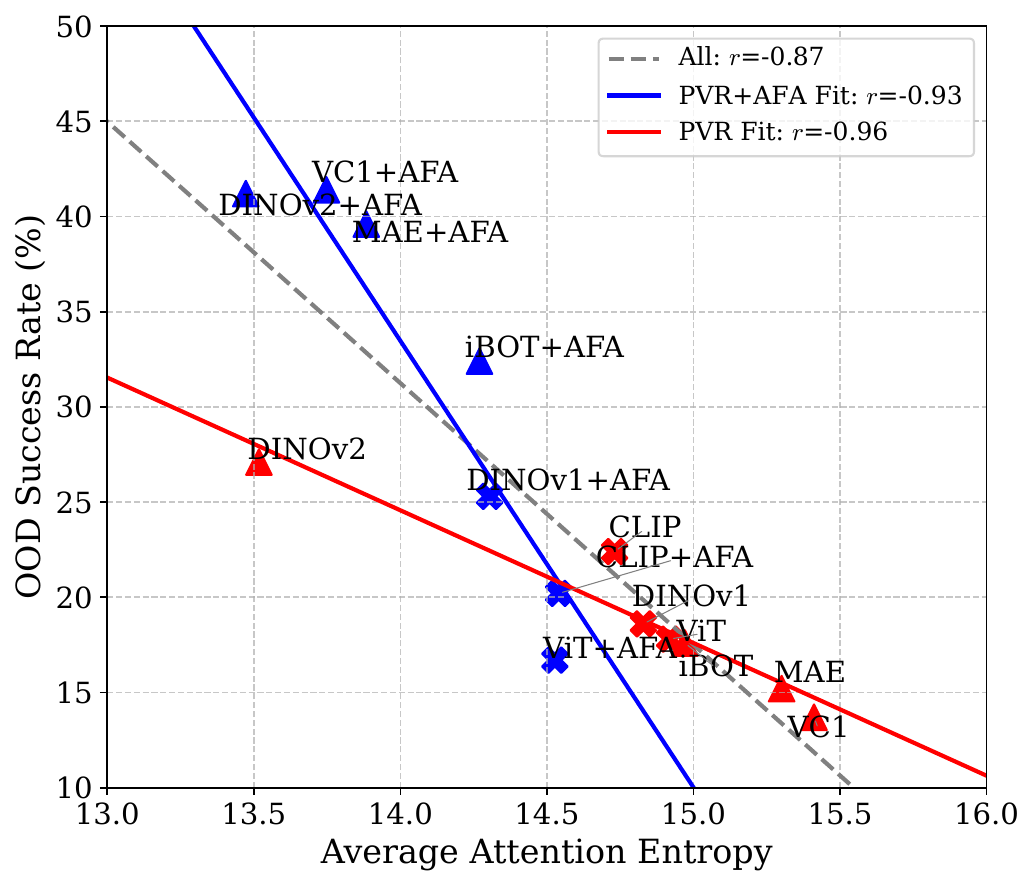}
    \end{subfigure}
    \caption{Correlation plots for OOD performance predictors. \textbf{Left}: Higher attention mass in task-relevant areas (e.g., robot, object) correlates with higher success. \textbf{Right}: Attention entropy negatively correlates with performance. \textbf{Legend}:~\textcolor{red}{\textbf{Red line (raw PVR)}},~\textcolor{blue}{\textbf{Blue (PVR+AFA)}},~\textcolor{gray}{\textbf{Gray (overall trend)}}.}
    \label{fig:correlations_sims}
    \vspace{-1em}
\end{figure*}
 
\subsection{Policy Robustness Evaluation}
\label{ssec:robustness_eval}
To evaluate robustness, we compare four distinct approaches both in ID and OOD scenarios, all trained without any domain randomisation. 
First, we train policies directly with global PVR features (annotated as PVR). 
The remaining three policies all aggregate local features using different methods: PVR+TL uses TokenLearner, PVR+SS employs Spatial SoftMax, and PVR+AFA utilizes our Attentive Feature Aggregation.
We summarise the performance of each policy category in Fig.~\ref{fig:results}, averaged across tasks and perturbation types in the OOD case. 
From these results, several trends emerge.

First, AFA yields the best OOD performance, in many cases even tripling the success rate (\eg VC-1, MAE, VIP, etc.). 
The only exceptions are some of the worse performing PVRs: CLIP, VIT and R3M. 
Regarding, the first two, no pooling method outperforms the raw PVR case.
This is reasonable, as these models are trained with objectives that emphasise global frame perception, unlike other models that incorporate supervision at the patch level. 
For the case of R3M, PVR+TL leads to slightly higher policy success rate. 
Notably, MIM-trained PVRs (\ie VC-1, DINOv2, MAE, and iBOT) benefit the most from AFA, reflecting the alignment between AFA's design, which is inspired by attentive probing, and the training principles of MIM-based models. 
These findings highlight AFA's ability to enhance policy performance, particularly in challenging OOD scenarios, and underscore its compatibility with models that leverage local feature representations.

Second, the average in-domain performance remains nearly unchanged between PVR and PVR+AFA, with a slight increase from $63.1$\% to $66.4$\% for the latter approach. 
The minor boost observed with AFA in in-domain scenarios, especially when compared to its substantial improvements in perturbed scenes, suggests that AFA does not learn a new latent space for the PVR more suited to the task. 
Instead, it appears to refine the use of the existing latent space by learning to leverage relevant information while discarding elements that are irrelevant to the policy. 
This distinction underscores AFA's role as a mechanism for better utilisation of pre-trained features rather than redefining or adapting the underlying feature space.
Furthermore, PVR+SS seems to lead to a slight boost in ID performance, compared to both PVR and PVR+AFA, but this performance is not retained in the OOD evaluation, where in all but three encoder choices (\ie VIP, DenseCL, and R3M) PVR+SS demonstrated notably degraded performance even to the raw PVR-trained policies.   
Finally, PVR+TL seems to perform poorly both in the ID and OOD cases.
We primarily attribute this to the fact that TL deploys global average pooling, which removes spatial information from the tokens.
TL was developed as a video classification module, thus being more fit for answering the question: \textit{``Is X in the scene?''}, rather than: \textit{``Where is X in the scene?''}, which is more important in visuomotor policy learning.  
It is also noteworthy that the original TL (and its adoption by RT-1) jointly trained the spatial attention mechanism with the image encoder, allowing the latter to adapt to the attention mechanism. 
In contrast, in the context of PVR-based policies, this flexibility is inherently restricted, limiting the compatibility of the PVR and TL modules.

Regarding the case of OOD scenes, we argue that the reduced performance of PVR+TL is also a consequence of how it derives attention. 
TL computes spatial attention weights strictly as a function of the input feature map, relying heavily on the in-domain statistics it was trained on. 
Unlike AFA, which uses a stable learned query to look for specific content, TL depends entirely on the current input's properties. 
Consequently, when scene statistics shift radically in OOD scenarios, these input-dependent weights become unstable and unpredictable.
Additionally, Spatial Softmax does not try to favour specific visual cues, but rather compresses the image information. 
As a result, redundant cues that are unrelated to the task remain.  

\begin{figure*}[t]
\vspace{1.5em} 
    \centering
   \includegraphics[clip, trim=0cm 0cm 0cm 1.5cm, width=0.98\textwidth]{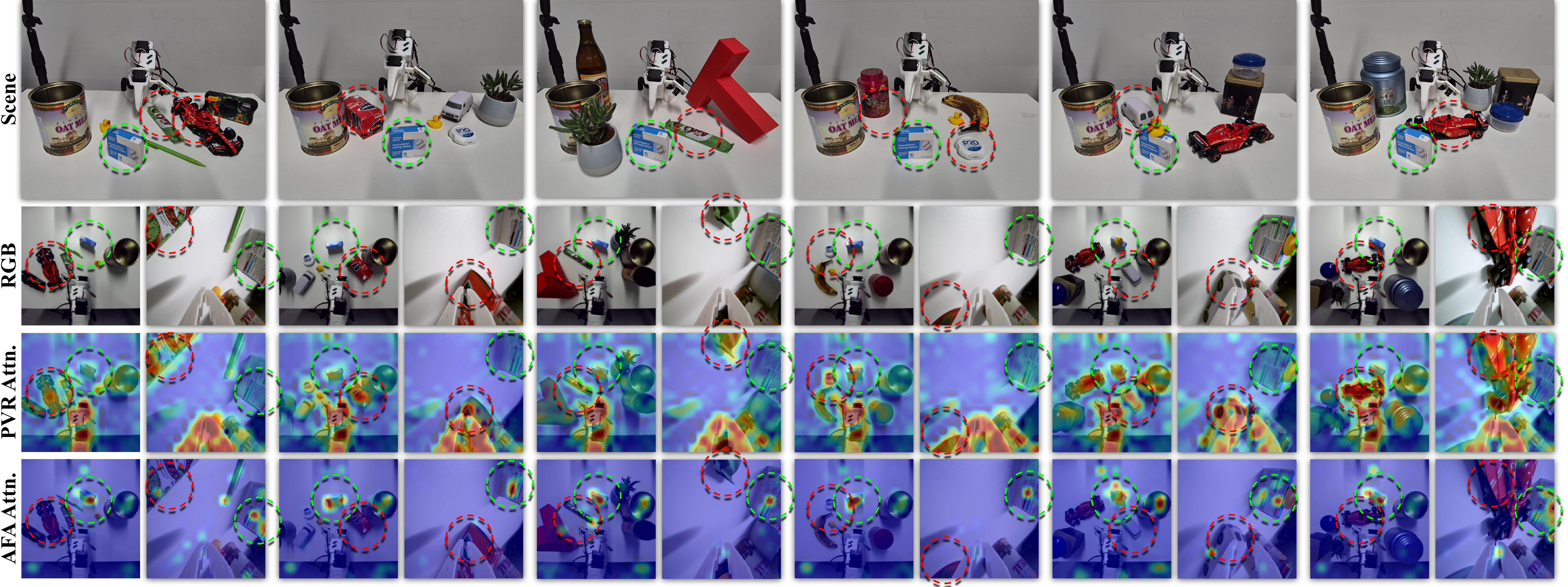}
    \caption{Visual attention comparison between PVR and PVR+AFA (ours). The top two rows show the initial setup with distractors and the robot camera views. The bottom rows show the respective attention heatmaps. We mark the object of interest with a \textcolor{green}{\textbf{green circle}} and strong distractors with a \textcolor{red}{\textbf{red circle}}. Videos are available on our~\texttt{\href{https://tsagkas.github.io/afa}{project page}}.}
    \label{fig:real-world}
    \vspace{-1em}
\end{figure*}

\subsection{OOD Performance Predictors}
\label{sec:ood_predictors}
We find two trends among PVRs that correlate with high OOD policy success rates and provide evidence that AFA enhances these attributes. 
All examined trends concern the attention maps extracted from the studied ViTs, utilising the provided expert demonstrations. 
First, we measured the average attention mass that falls within the robot arm and objects of interaction.
For this, we extracted the masks using SAMv2~\cite{ravi2024sam2}.
We find very strong positive correlation ($\rho=0.81$) between this metric and the OOD performance and observe that AFA magnifies this behaviour (see left plot of Fig.~\ref{fig:correlations_sims}). 
Second, we measure how much targetted the attention is, by computing the entropy of the attention heatmap. 
We find very strong negative correlation ($\rho=-0.87$) between the entropy and the OOD success rate (see right plot of Fig.~\ref{fig:correlations_sims}).
Similarly to before, AFA enhances this behaviour, as visualised in Fig.~\ref{fig:teaser} and Fig.~\ref{fig:real-world}, where attention appears to be extremely focused.
Consequently, AFA seems to successfully learn to attend to task-relevant visual cues, ignoring changes in the scene (\eg semantically rich distractor objects). 

\subsection{Real-World Experiments}
\label{sec:real_world}
We also validated the efficacy of AFA in the real-world using two different robot platforms and tasks. 
Initially, we conducted experiments using the LeRobot SO-101 robot platform~\cite{cadene2024lerobot}, utilising the provided wrist camera and an additional external ZED2i from a top-down view. 
The chosen task involved picking up a small  blue cardboard box and placing it inside of a cylindrical tin can.
During evaluation, we awarded $0.5$ a point for successfully picking up the object and another $0.5$ for placing it in the can.
The object's initial pose was randomised within a 90$^\circ$ arc at a 30 cm radius from the robot's base, whereas the container remained stationary.
In total, $60$ demonstrations were collected, without any scene distractors, and the policy was trained with and without AFA. 
In order to further emphasise the lasting value of our findings, we run our experiments with DINOv3~\cite{simeoni2025dinov3} PVR features, which was released after we conducted our simulation experiments. 
We evaluated the trained policy both ID and under visual perturbations, in the presence of task-irrelevant ``distractor'' objects.
Overall, $20$ everyday items were utilised as distractors, with up to 7 of them being randomly placed in the scene at the beginning of each OOD rollout.

The success rate of each (policy, domain) pair was measured over a total of $20$ different rollouts, is reported in Table~\ref{tab:real_world}. 
Even though the ID performance between PVR and PVR+AFA is comparable (87.5\% and 85.0\%, respectively), in the OOD setting, the success rate of the policy trained without AFA plummeted to 17.5\% (failing to complete the task even when a single distractor object is present in the scene). 
AFA only resulted in a reactively minor performance drop, achieving a success rate of 75\%. 
We visualise an analysis of these results in Fig.~\ref{fig:real-world} where we observe that AFA manages to focus attention mostly to task-relevant regions, whereas the PVR's attention is distributed across every semantically rich object in the scene.

\begin{figure}[t]
\centering
\includegraphics[width=1\linewidth]{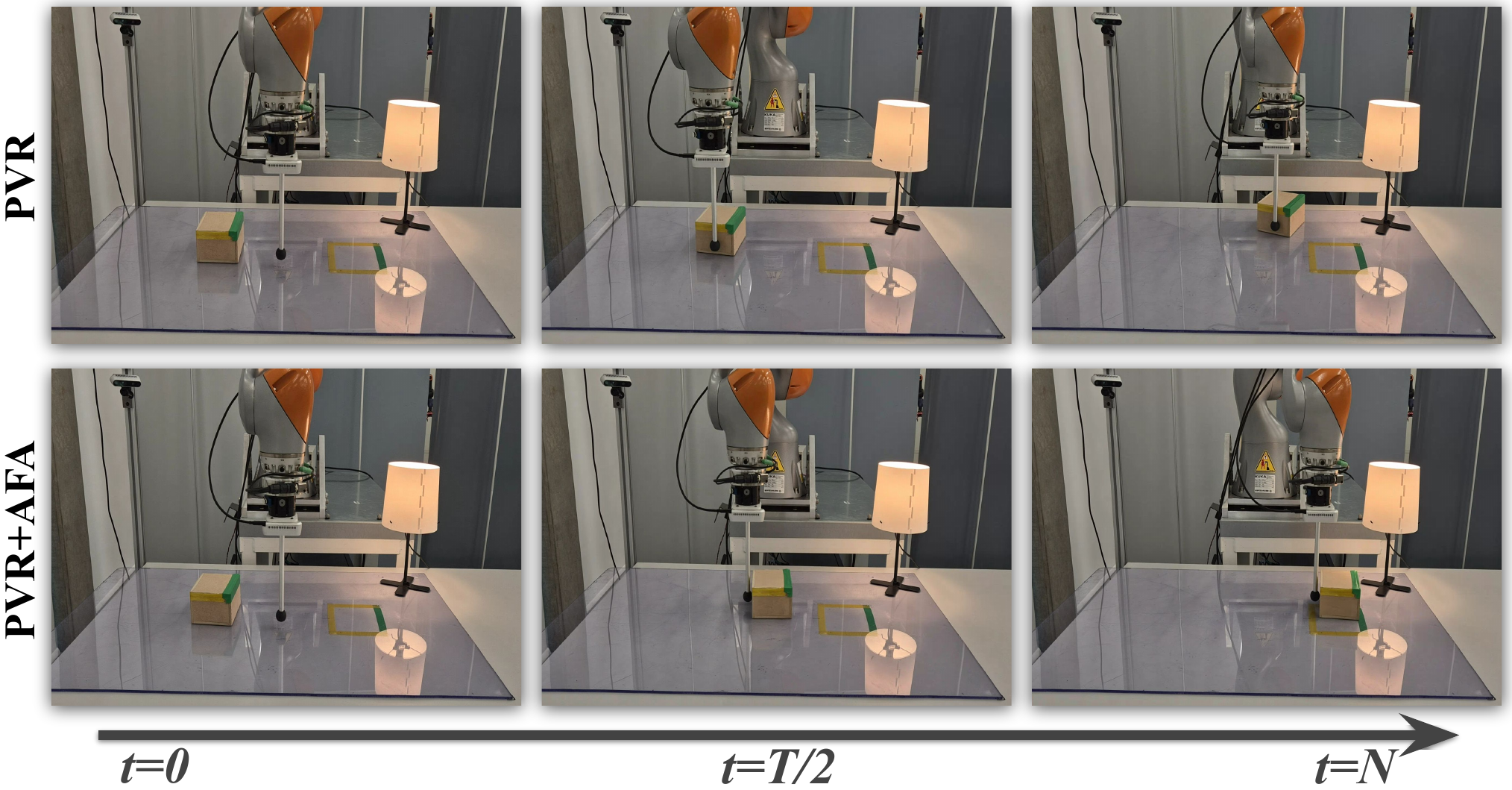}
\caption{Policy rollout example under scene perturbations.} 
\label{fig:kuka_lighting}
\vspace{-2em}  
\end{figure}

We also conducted smaller-scale experiments (10 ID and 16 OOD rollouts in total) on a planar-pushing task, using a KUKA IIWA 14. 
As listed in Table~\ref{tab:real_world}, due to the task's contact-rich nature, without AFA, the PVR-based policy to failed catastrophically in all OOD rollouts (with success rate dropping to $0\%$) under both distractor additions and lighting changes. Notably, the vanilla PVR-based policy consistently demonstrated a miss-oriented behaviour attributed to the OOD drift of its input, representatively demonstrated in the example rollout of Fig.~\ref{fig:kuka_lighting}. 
Conversely, with AFA, the policy maintained the ID performance across rollouts, indicating its effectiveness in rejecting task-irrelevant visual cues, maintaining the policy input in domain. 
All real-world experiments are available on our~\texttt{\href{https://tsagkas.github.io/afa}{project page}}.

\begin{table}[t]
    \centering
    \caption{Real-world policy rollout success rate.}
    \label{tab:real_world}
    
    \newcolumntype{Y}{>{\centering\arraybackslash}X}
    
    \begin{tabularx}{\linewidth}{l|YY|YY}
        \toprule
        & \multicolumn{2}{c|}{\textbf{Pick and Place}} & \multicolumn{2}{c}{\textbf{Planar Pushing}} \\
        & {ID} & {OOD} & {ID} & {OOD} \\
        \midrule
        PVR       & 87.5\% & 17.5\% & 80\% & 0\%   \\
        PVR + AFA & 85.0\% & 75.0\% & 80\% & 100\% \\
        \bottomrule
    \end{tabularx}
    \vspace{-2em} 
\end{table}
\section{DISCUSSION}
We demonstrated that the robustness of visuomotor policies leveraging PVRs is critically dependent not just on the encoder itself, but on how its features are aggregated. 
Our AFA approach offers a simple yet powerful method to learn to focus on task-relevant visual cues, significantly improving OOD performance without the need for costly dataset augmentation or PVR fine-tuning. 
This finding suggests a promising path towards more robust robotic systems that can operate reliably in visually dynamic environments. 

Nevertheless, a question naturally arises:
\textit{``which PVR characteristics are most suitable for robot learning?''}, particularly in OOD scenes.
From the selected PVRs, VIP, VC-1, and R3M were developed with visuomotor policy learning as the target application. However, this does not seem to be a decisive factor for high success rate and robustness. Similarly, whether the backbone of the PVR is a CNN or a vision transformer seems to be playing a small role.  
In-domain, VFS and R3M score the highest performance (close to 80\%), but this is most likely due to their temporal perception~\cite{tsagkas2025temporaltrapentanglementpretrained}.
On the other hand, OOD performance seems to depend mostly on the MIM-training strategy when combined with AFA, and not on the training dataset (\eg if it's static or from video).
This is not unreasonable, given the powerful local perception that this methodology leads to.
Similarly, these observations highlight that we are still to uncover an optimal approach for building vision encoders for policy learning.








\bibliographystyle{abbrv} 
\bibliography{references}


\end{document}